\documentclass[lettersize,journal]{IEEEtran}
\usepackage{amsmath,amsfonts}
\usepackage{algorithmic}
\usepackage{algorithm}
\usepackage{array}
\usepackage[caption=false,font=footnotesize,labelformat=simple]{subfig}

\usepackage{textcomp}
\usepackage{stfloats}
\usepackage{url}
\usepackage{verbatim}
\usepackage{graphicx}
\usepackage{cite}
\usepackage{float}
\usepackage{comment}
\usepackage{nomencl}
\usepackage{etoolbox}
\usepackage{multirow}
\usepackage{booktabs}
\usepackage{gensymb}
\usepackage[colorlinks=true]{hyperref}
\begin{document}

\title{Circular Accessible Depth: A Robust Traversability Representation for UGV Navigation}

\author{
    Shikuan~Xie, 
    Ran~Song, 
    Yuenan~Zhao, 
    Xueqin~Huang, 
    Yibin~Li,
    and Wei~Zhang*

\thanks{All authors are with the School of Control Science and Engineering, Shandong University, Jinan, China.}
\thanks{X. Huang is also with the Fuxi AI Lab, NetEase Games, Hangzhou, China.}
\thanks{*Corresponding author: Wei Zhang (Email: davidzhang@sdu.edu.cn)}
}



\maketitle

\begin{abstract}
In this paper, we present the Circular Accessible Depth (CAD), a robust traversability representation for an unmanned ground vehicle (UGV) to learn traversability in various scenarios containing irregular obstacles.
To predict CAD, we propose a neural network, namely CADNet, with an attention-based multi-frame point cloud fusion module, Stability-Attention Module (SAM), to encode the spatial features from point clouds captured by LiDAR.
CAD is designed based on the polar coordinate system and focuses on predicting the border of traversable area. 
Since it encodes the spatial information of the surrounding environment, which enables a semi-supervised learning for the CADNet, and thus desirably avoids annotating a large amount of data.
Extensive experiments demonstrate that CAD outperforms baselines in terms of robustness and precision.
We also implement our method on a real UGV and show that it performs well in real-world scenarios.
\end{abstract}


\section{Introduction}

\IEEEPARstart{M}{ost} existing work \cite{Survey, Survey2} on autonomous navigation focuses on urban self-driving, where only pre-defined and/or structured objects, such as vehicles and pedestrians, are recognized via 3D detection techniques \cite{MV3D, VoxelNet, SECOND, PIXOR, PointPillars}.
Such perception approaches may cause serious accidents due to the lack of a robust traversability prediction in the case that some objects do not present in the training dataset or have irregular shapes that cannot be represented by regular 3D boxes. In particular, this problem becomes more pronounced for unmanned ground vehicles (UGV), which inevitably confront irregular obstacles such as trunks, lamp posts, barriers, and even downward stairs when performing searching, inspection, and delivery tasks in environments significantly different from a typical highway for autonomous driving.


Bird's-eye view (BEV) semantic map has recently received increasing attention \cite{semantic_inpaint, SSC-IROS2021, semantic_CoRL, MASS, BEV-Seg, CrossView-RAL2020} for traversability representation in autonomous navigation.
This kind of representation treats traversability prediction as a semantic segmentation of the surrounding environment.
On the one hand, it provides richer and more diverse information on road traversability than most 3D detection techniques, which improves the security of autonomous navigation. On the other hand, it describes the traversability uniformly in 2D BEV, which facilitates the subsequent planning and control for UGV. However, BEV semantic map has two major issues which significantly affect the robustness of the traversability prediction.

First, BEV semantic map is not robust in predicting thin objects, such as pedestrians, poles, and barriers.
The root cause of this issue is that a BEV semantic map has no spatial constraint on the border of traversable area.
This is because it is essentially a pixel-wise classification which focuses only on the semantic features of different locations but not on the spatial distribution of the border of the traversable area.
It is noteworthy that the commonly used cross-entropy loss does not impose any spatial constraint on the border of the traversable area, which leads to an imprecise prediction of the border location. Also, since some thin objects, such as pedestrians, lamp posts, and road barriers, occupy only a small number of pixels in the BEV, the constraint of pixel-wise classification alone cannot compel the BEV network extract informative features for such obstacles, leading to an unreliable prediction for thin obstacles.

Second, BEV semantic map is not robust in predicting dynamic objects, such as moving cars and bicycles, due to the generation process of the ground truth of BEV semantic map.
Peng et al. \cite{MASS} demonstrated that dense ground truth is important for the prediction of BEV semantic map, and thus densification is an essential part of the ground truth production. However, the LiDAR point clouds are usually sparse. Most methods generate desne ground truth by concatenating multi-frame point clouds. As a result, moving objects may leave undesired trailing shadows in the ground truth.
To address this problem, Shaban et al. \cite{semantic_CoRL} and Peng et al. \cite{MASS} proposed to keep only single-frame point clouds of moving objects for generating the ground truth. 
This tactic results in the objects of the same category being annotated with different labels, leading to unreliable prediction of dynamic objects.

Aiming at the above two issues, we propose a robust LiDAR-based traversability representation for UGV navigation, called Circular Accessible Depth (CAD).
Same as BEV semantic map, CAD can uniformly represent different kinds of obstacles for predicting the traversability of the surrounding environment.
However, instead of predicting the semantic category for each pixel-wise location, CAD predicts the maximum accessible depth in all circular directions centered at the UGV.
For the first issue, unlike BEV semantic map based on the Cartesian coordinate system, CAD is based on the polar coordinate system and thus directly expresses the distance to the border of the traversable area, which facilitates the introduction of the spatial constraint.
With the representation based on the polar coordinate system and the spatial constraint, CAD predicts the border of the traversable area more robustly. 
Benefiting from the polar representation, thin objects occupy more directions at closer distances, which further improves the robustness of the prediction of thin objects.
For the second issue, CAD focuses only on predicting the locations of obstacles rather than their shapes in BEV.
Therefore, the ground truth of CAD does not suffer from the inconsistent annotations of dynamic and static objects that exists in a BEV semantic map. With more reliable annotations, CAD provides more robust predictions for dynamic objects.

We present a neural network, namely CADNet, based on the CAD representation to analyze road traversability for UGV navigation by learning the features with regard to the spatial distribution of LiDAR point clouds.
Different from BEV methods which predict the independent probabilities of semantic categories, CADNet predicts the probability distribution of the maximum accessible depth along the radial direction in the spatial dimension.
Although the training of CADNet is supervised, benefiting from the inherent property of probability distribution of the CAD representation, we are able to introduce a semi-supervised training to restrict both the entropy and the variance of the probability distribution to make it more concentrated in the spatial dimension.
As such, we only need to annotate a small number of labels to inform CADNet what kinds of points potentially represent obstacles or border of traversable area for UGV navigation. Also, the semi-supervision enables the learning of a wider sample distribution from a large amount of unlabeled data.
Since the ground truth annotations for CAD are easier to be generated than those for BEV semantic map, its applicability to semi-supervised learning ensures a high efficiency in real-world applications.

The traversability prediction faces the problems of large LiDAR blind spots and the sparsity of point clouds.
Single-frame point cloud lacks sufficient information for the prediction of the traversability, especially facing low and negative obstacles, and thus it is important to fuse historical information to assist in the prediction.
However, dynamic objects such as moving vehicles and pedestrians may leave trailing shadows in the historical points, which seriously interferes with the traversability prediction.
Since the BEV semantic map contains semantic information of dynamic objects, 
Shaban et al. \cite{semantic_CoRL} proposed to eliminate the dynamic features remaining in the memory via an RNN module.
In contrast, CAD does not include semantic information about dynamic objects, making it inapplicable to the RNN approach.
To cope with this problem, we start from the perspective of feature stability and propose an attention-based multi-frame point clouds fusion module, namely Stability-Attention Module (SAM).
SAM calculates the correlation coefficients between the features of each frame and the features of the whole sequence as the attention weights.
The features of dynamic objects are unstable and have less correlation with the overall features of the whole sequence, and thus will be assigned smaller attention weights and eliminated in the final fused features.
This significantly reduces the interference of historical points of dynamic objects in CAD prediction.

We also develop an easy-to-use CAD annotation tool to facilitate the application of our method in the real world.
We demonstrate the feasibility of the proposed CAD in practical UGV navigation experiments by combining some commonly used path planning methods.
Extensive experiments show that CAD outperforms baselines in terms of the robustness of the traversability prediction.

Overall, the contributions of this paper are threefold:
\begin{itemize}
\item{We propose Circular Accessible Depth, a robust traversability representation for UGV navigation.}
\item{We present the CADNet, trained in a semi-supervised manner, to extract the features with regard to the spatial distribution of point clouds and predict CAD.}
\item{In the CADNet, we design a Stability-Attention Module to tackle the interference of dynamic objects in multi-frame point clouds fusion.}
\end{itemize}

\section{Related Work}

This section first reviews different types of traversability representation for UGV navigation.
Then, it briefly reviews some relevant LiDAR point processing methods, including multi-frame point clouds fusion methods necessary for traversability prediction.

\subsection{Traversability Representation for UGV Navigation}

Usually, UGV navigation is not supported with high-definition maps and confronts diverse obstacles.
Thus, designing a flexible and efficient traversability representation subject to the surrounding environment is crucial.
Some existing methods \cite{STVL, neg_IROS, neg_RAL} attempted to model the surrounding environment as a 3D occupancy map according to the LiDAR point clouds to calculate the traversability.
However, such methods lead to high memory consumption and computational costs, especially in vast outdoor scenarios, making it difficult for the UGV to operate rapidly and flexibly.
Recently, BEV semantic map has received more and more attention \cite{semantic_inpaint, SSC-IROS2021, semantic_CoRL, MASS, BEV-Seg, CrossView-RAL2020} due to the the rich traversability information it provides.
\cite{semantic_inpaint, SSC-IROS2021} considered it as a scene completion task to complete sparse semantic LiDAR points into semantic map via inpainting.
\cite{semantic_CoRL, MASS, BEV-Seg, CrossView-RAL2020} proposed to predict the BEV semantic map directly from LiDAR points or camera images.
However, such semantic segmentation approaches consider only the semantic classification of each location, but lack the constraints in the spatial dimension of the edges of the traversable area.
There also exists some traditional approaches \cite{PointCloud2LaserScan, Neg_JFR, Neg_ITSC} to calculate the obstacle distance by the geometry of point clouds.
For example, Bovbel et al. \cite{PointCloud2LaserScan} converted points within a height range to a laser scan, which can only handle obstacles of a certain height.
Shang et al. \cite{Neg_JFR} and Larson et al. \cite{Neg_ITSC} focused on detecting negative obstacles while had strict requirements for LiDAR setup, limiting their applications.
Instead, we propose CAD to represent the traversability and present the CADNet to learn the spatial distribution of LiDAR points for a wider range of applications.

\subsection{LiDAR Points Processing for UGV Navigation}

A key role of the perception module in LiDAR-based UGV navigation is to extract spacial features from the point clouds.
Early methods \cite{voting, Vote3Deep} for 3D object detection often used voxel grid instead of points to represent spatial features.
After PointNet \cite{PointNet} was proposed, some methods such as VoxelNet \cite{VoxelNet}, SECOND \cite{SECOND}, and PointPillars \cite{PointPillars} proposed to first divide the space into partitions and encode the local features for each partition by PointNet, and then extract the global features by CNN.
While most methods divide the point clouds under a Cartesian coordinate system, Zhu et al. \cite{Cylinder3D} and Zhang et al. \cite{PolarNet} proposed to describe the spatial features in polar coordinate system to equalize the distribution of points in each partition.
Considering that CAD is represented in a circular form, to keep the geometric unity of the spatial features, we designed the perception module in the polar coordinate form.


Due to the sparsity of LiDAR point clouds, it is difficult to obtain sufficient spacial information from a single-frame point cloud.
Thus, multi-frame point clouds fusion is an important part for traversability prediction.
Most of the existing multi-frame point clouds fusion methods were proposed for 3D detection \cite{offboard, LSTM3DDet, wysiwyg, MP3}, where only foreground points were fused.
However, such methods are not applicable for the task of traversability prediction where the LiDAR points are not classified as foreground or background.
Since CAD does not contain semantic information, RNN-based method \cite{semantic_CoRL} used for BEV semantic map is also not applicable.
Some other approaches \cite{semantic_CoRL, wysiwyg, Costmap2D} attempted to eliminate the effects of dynamic objects via ray tracing \cite{RayTracing}.
One disadvantage of such methods is that it is computationally intensive, while a more serious problem is that for many outdoor environments, LiDAR may not be able to produce effective laser echoes, and thus ray tracing cannot be calculated.
Instead, we start from the perspective of feature stability and propose an attention-based multi-frame point clouds fusion module to keep only the stable features.

\begin{figure*}[t]
    \centering
    \includegraphics[width=1.0\textwidth]{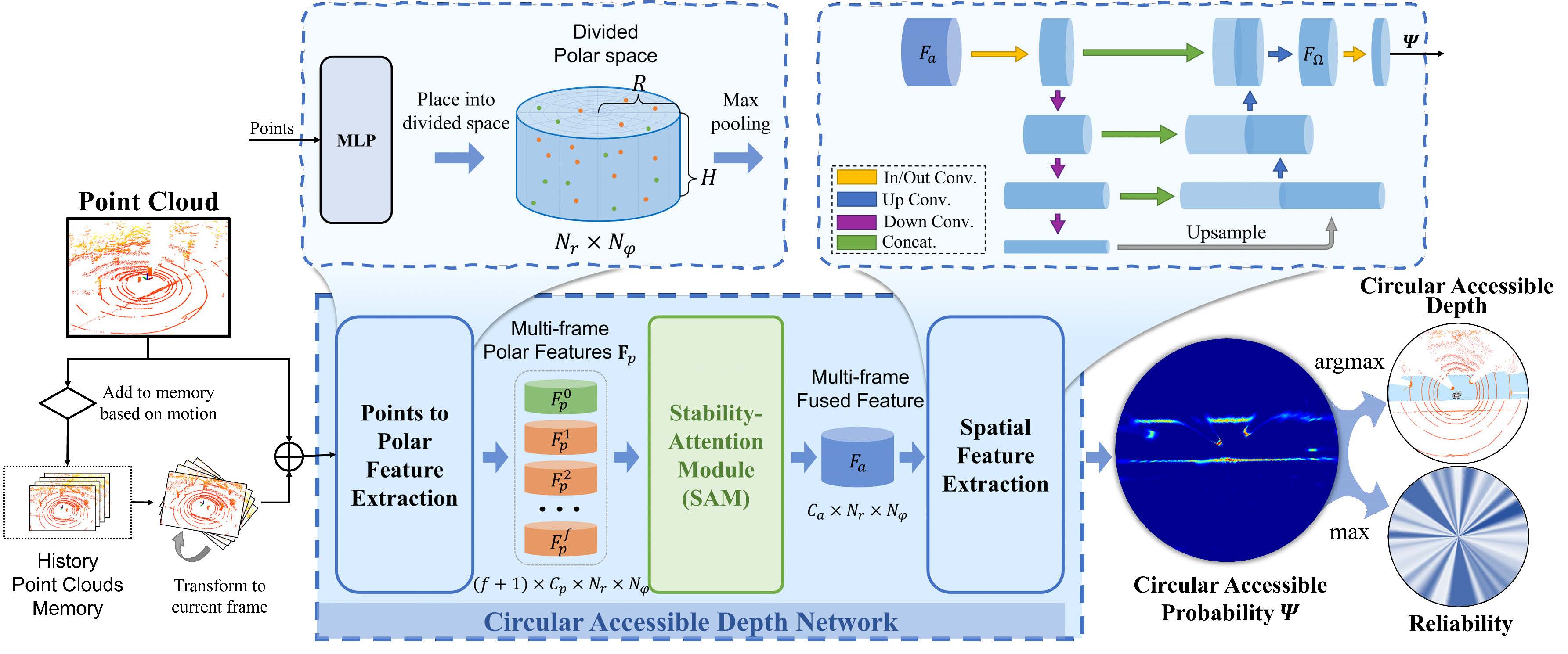}
    \caption{The pipeline of CADNet including three main components \textit{Points to Polar Feature Extraction}, \textit{Stability-Attention Module (SAM)}, and \textit{Spatial Feature Extraction.}}
    \label{fig:pipeline}
\end{figure*}

\section{Methods}

This section consists of three parts.
First, we introduce the network structure of CADNet and the processing flow of the point clouds. Then, we analyze in detail the Stability-Attention Module of CADNet for multi-frame fusion.
Finally, we elaborate the semi-supervised training process and the loss functions for CADNet.
The key symbols in this section are denominated as listed in Table \ref{tab:Nomenclature}.

\begin{table}[htbp]
    \centering
    \caption{Nomenclature\label{tab:Nomenclature}}
    \begin{tabular}{|p{0.1\textwidth}|p{0.32\textwidth}|}
        \hline
        \textbf{Symbols} & \multicolumn{1}{c|}{\textbf{Description}} \\
        \hline
        \multicolumn{2}{|c|}{\textbf{CADNet}} \\
        \hline
        $N_l$, $N_u$ & Numbers of labeled samples and unlabeled samples \\
        \hline
        $N$ & Dataset size, satisfying $N=N_l+N_u$ \\
        \hline
        $N_r$, $N_\varphi$ & Number of partition along $r$-axis and $\varphi$-axis \\
        \hline
        $f$ & Number of historical frames \\
        \hline
        \multirow{2}[0]{*}{$i$, $j$, $k$} & Indices for data samples, directions of CAD, and frames \\
        \hline
        $S_{ik}$ & A set of points \\
        \hline
        $N_{ik}$ & Number of points in $S_{ik}$ \\
        \hline
        $p_{ik}$ & Pose of LiDAR \\
        \hline
        $L_i$ & Label of the $i$-th sample \\
        \hline
        \multirow{2}[0]{*}{$l_{ij}$} & Index formed depth annotation of the $j$-th direction in $L_i$ \\
        \hline
        $\mathbf{F}_p$, $F_h$, $F_a$ & Polar features, historical feature, and fused feature \\
        \hline
        $C_p$, $C_a$ & Number of channels of $\mathbf{F}_p$ and $F_a$ \\
        \hline
        \multicolumn{2}{|c|}{\textbf{Semi-supervised Training}} \\
        \hline
        $\Psi^l_i$, $\Psi^u_i$ & Predicted probability for labeled/unlabeled samples \\
        \hline
        $Y_i$ & One-hot coding form of the label $L_i$ \\
        \hline
        $t$ & Training epoch \\
        \hline
        $d$ & Index of partition along $r$-axis \\
        \hline
        \multirow{2}[0]{*}{$y_{jd}$, $\psi_{jd}^l$, $\psi_{jd}^u$} & The $d$-th probability value of the $j$-th direction in $Y$/$\Psi^l$/$\Psi^u$ \\
        \hline
        $\alpha$, $\beta$, $\lambda$ & Balance factors in $\mathcal{L}_{l}$, $\mathcal{L}_{u}$ and $\mathcal{L}$ \\
        \hline
        $g$ & Distance weight \\
        \hline
        $b$, $m$ & Number/Index of adjacent partition used in $\mathcal{L}_{reg}$ \\
        \hline
    \end{tabular}
\end{table}

\subsection{CADNet}
CADNet aims to predict the CAD representation of the surrounding environment by extracting the spatial feature of LiDAR point clouds.
Since our work is to predict the traversability of the surrounding environment centered on the UGV, the common orthogonal meshing of the point clouds conducted in the Cartesian coordinate system destroys the continuity of the UGV-centered spatial description.
By comparison, predicting the probability distribution of the traversability for continuous space along the radial axis in the polar coordinates is more suitable for our task.
Thus, the spatial representation and the spatial feature extraction structure of the network are both consistent with the polar coordinate system.

As shown in Fig. \ref{fig:pipeline}, CADNet contains three main components: \textit{Points to Polar Feature Extraction}, \textit{Stability-Attention Module (SAM)}, and \textit{Spatial Feature Extraction}, where SAM will be discussed in detail in the next subsection.

\subsubsection{Problem Statement}
A training dataset $N = N_l + N_u$ includes $N_l$ labeled samples $\{(S_{ik}, p_{ik}, L_i)|i=1, ..., N_l, k=0, ..., f\}$, and $N_u$ unlabeled samples $\{(S_{ik}, p_{ik})|i=1, ..., N_u, k=0, ..., f\}$, where $S_{ik} \in \mathbb{R}^{N_{ik}\times4}$ is the $k$-th frame in the $i$-th point cloud and contains $N_{ik}$ points.
A point cloud consists of $f+1$ frames, including one current frame and $f$ historical frames with totally $\sum_{k=0}^{f} N_{ik}$ points.
Each row of $S_{ik}$ represents a LiDAR point with four elements including the three-dimensional coordinates of the point and its LiDAR echo intensity.
$p_{ik}$ is the pose of the LiDAR sensor when the point cloud $S_{ik}$ is captured.
$L_i$ is the CAD label corresponding to the point cloud $S_{i0}$ of the current frame, expressed as $L_i = \{l_{ij}|j=1, ..., N_\varphi\}$, where $N_\varphi$ indicates that each label contains the accessible depth for $N_\varphi$ directions and $l_{ij}$ is the depth in the index form for the $j$-th direction.
The goal of the perception module is to learn a model to minimize the mean absolute error between the prediction and the label.

\subsubsection{Points to Polar Feature Extraction}
The space around the robot, a cylinder centered on the LiDAR with radius $R$ and height $H$, is divided into $N_r \times N_\varphi$ sector pillars as shown in the \textit{Points to Polar Feature Extraction} part in Fig. \ref{fig:pipeline}, where $N_r$ and $N_\varphi$ denote the number of divisions along the $r$-axis and the $\varphi$-axis respectively.
The division of the space also determines the shape of the spatial features.
The historical points $S_{ik}, k=1, ..., f$ are first transformed into the current coordinate system subject to the transformation $T_{ik}$ between the historical pose $p_{ik}$ and the current pose $p_{i0}$.
Then all the points are input into the MLP-Maxpool structure proposed by PointNet \cite{PointNet} to construct local spatial features with cylindrical shape.
It is noteworthy that since the points belong to different frames, the max pooling should be applied to the points which belong to the same frame and are located in the same sector pillar.
Thus, the MLP-Maxpool structure creates $f+1$ polar features $F_p^k, k=0,1,...,f$ for $f+1$ frames.
And the overall polar features can be expressed as $\mathbf{F}_p = \{F_p^k | k=0,1,...,f\}$, whose shape is $(f+1, C_p, N_r, N_\varphi)$, where $C_p$ is the number of channels.

Then, $\mathbf{F}_p$ is input to the Stability-Attention Module to be aggregated as multi-frame fused feature $F_a$.

\subsubsection{Spatial Feature Extracting}
Whether it is the polar feature extracted by the MLP-Maxpool structure or the multi-frame fused feature aggregated by SAM, they are all local features with respect to the location where each sector pillar is located.
To obtain the final CAD, the features of the whole space need to be encoded.
Analogous to \cite{PointPillars, PolarNet}, we adopt a UNet \cite{UNet}-like encoder-decoder structure, as shown in the \textit{Spatial Feature Extraction} part in Fig. \ref{fig:pipeline} to encode the adjacent spatial features.

Since the output of our method is a probability distribution, we finally use \texttt{argmax} to generate the CAD representation, and also output this max probability value as the prediction reliability in this direction to the downstream planning module as a reference.

\begin{figure*}[t]
    \centering
    \includegraphics[width=1\textwidth]{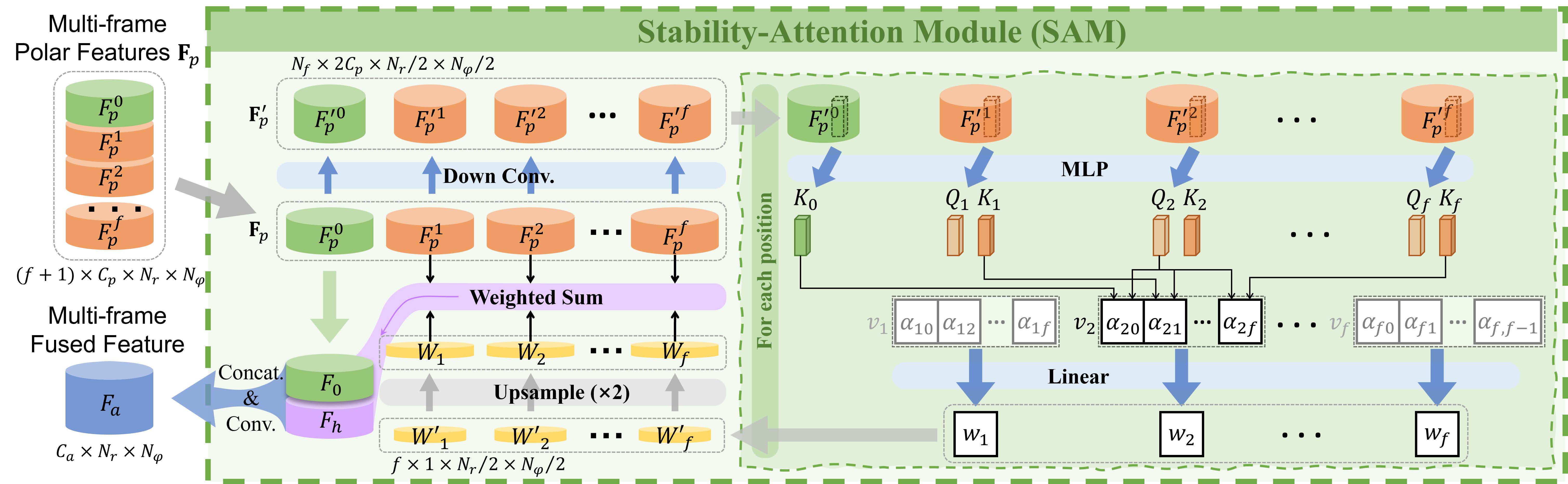}
    \caption{Stability-Attention Module (SAM). SAM eliminates the unstable dynamic features in historical features to fuses multi-frame point clouds. The right part of this illustration describes in detail the computing process for each position of different frames, where the computation of correlation vector takes the frame $t-2$ as an example.}
    \label{fig:SAM}
\end{figure*}

\subsection{Stability-Attention Module}
\label{SAM}

\subsubsection{Dynamic Feature Analysis}

Some existing methods utilize point-wise \cite{offboard, LSTM3DDet, wysiwyg} or pixel-wise \cite{semantic_CoRL} semantic information to cope with the interference of dynamic objects in multi-frame fusion.
However, CADNet cannot obtain semantic labels from the CAD ground truth for which points belong to dynamic objects.
Thus, it is required for CAD to eliminate the points of dynamic objects in the historical frames without knowing the semantic information of the points. We think about this problem from the perspective of feature stability.

As an example, a ground should be detected by LiDAR as a low and flat point cloud, which we call a ground feature.
If a pedestrian passes through the ground at a certain moment, then for this location, the spatial features at this moment are represented as a pedestrian feature.
For the overall point cloud sequence at this location, the pedestrian feature is transient, while the ground feature is stable.
In this way, how to eliminate the dynamic features depends on how to evaluate the feature stability.
We consider the feature stability as the correlation between the feature at this moment and the feature of the whole sequence. Then, a low correlation indicates that this feature is unstable and likely to be the feature of a dynamic object.
Hence, to fuse multi-frame point clouds for CAD, we design an attention-based module, which calculates the correlation between features.

\subsubsection{SAM Structure}

The architecture of SAM is illustrated in detail in Fig. \ref{fig:SAM}.
The polar features $F_p^k, k=0, 1, ..., f$ in the multi-frame polar features $\textbf{F}_p$ are all local features. Due to the sparsity of the point cloud, many locations have empty features, which is not conducive to correlation calculation.
Therefore, we first downsample $\textbf{F}_p$ to obtain $\textbf{F}_p'$ with more evenly distributed features.

For the polar feature $F_p^{'k}$ of each frame, we can obtain two embedded vectors, $Q_k$ and $K_k$, through MLP.
It is worth noting that since the point cloud of the current frame is absolutely accurate, the polar feature $F_p^{'0}$ related to the current frame is only used to be referred by other frames.
Thus, $F_p^{'0}$ is only embedded to \texttt{key} $K_0$ but not \texttt{query} $Q_0$.
Then, for each historical frame, $Q_k$ is multiplied by $K$ of the other frames to obtain a correlation vector $v_k$, expressed as:
\begin{equation}
\begin{aligned}
    v_k = \{\alpha_{kn} | n = 0, ..., f \land n \neq k\},
    \alpha_{kn} = Q_k \cdot K_n.
\end{aligned}
\end{equation}
Since it is meaningless to calculate its own correlation, $Q_k$ is not multiplied by $K_k$, and then the length of the correlation vector $v_k$ is $f$, while the length of the multi-frame feature sequence is $f+1$.
Then, the final attention weight $w_k$ is calculated by a learnable linear layer subject to the correlation vector $v_k$.
The weight $w_k$ represents the stability of the features of the $k$-th frame.
The more stable the feature is, the greater the value of the weight. Thus, the feature of dynamic objects is effectively removed from the historical features.

$w_k$ is the weight for one location, and the total weights of $F_p^{'k}$ can be represented as $W_k'$, whose shape is $(1 \times N_r/2 \times N_\varphi/2)$.
Since we downsample the multi-frame polar features $F_p^k$ at the beginning, the attention weights $W_k'$ have to be upsampled to $W_k$, whose size is the same as that of $F_p^k$.
Then, the historical polar features $F_p^k, k=1, ..., f$ are weighted by $W_k, k=1, ..., f$ and summed up to generate the fused historical feature $F_h$:
\begin{equation}
    F_h = \sum_{k=1}^{f} F_p^k \cdot W_k.
\end{equation}
As mentioned above, to ensure the completeness of the current feature $F_p^0$, instead of fusing them directly by a weighted summation, we use a convolutional layer to fuse $F_p^0$ with $F_h$:
\begin{equation}
    F_a = \mathrm{conv}(\mathrm{concatnate}(F_p^0, F_h)).
\end{equation}
$F_a$ is the final multi-frame fused feature and will be output to the subsequent module for spatial feature extraction.

\subsection{Semi-supervised Training}

Although the CAD prediction is supposed to be a regression task, in our experiments we found it extremely difficult for the network to directly regress a distance variable in a large continuous space.
Therefore, we regard it as an $N_r$-classification task, dividing each direction into $N_r$ parts, and predicting the probability distribution of the space to which the accessible depth belongs.

Since our network predicts the probability distribution of the maximum accessible depth along the radial direction on space dimension, we constrain the overall probability distribution to improve the effectiveness of semi-supervised learning.
The semi-supervised loss function is expressed as:
\begin{equation}
\begin{aligned}
    \mathcal{L}(S_\Psi^l, S_Y, S_\Psi^u, t)
        &= \frac{1}{N_l} \sum_{i=1}^{N_l}
            \mathcal{L}_{l}(\Psi_i^l, Y_i, t) \\
        &+ \lambda \frac{1}{N_u} \sum_{i=1}^{N_u}
            \mathcal{L}_{u}(\Psi_i^u, t)
    ,
\end{aligned}
\end{equation}
where $\Psi_i^l$ and $\Psi_i^u$ represent the prediction probability of the $i$-th labeled sample and unlabeled sample respectively. $Y_i$ is the one-hot coding form of the label $L_i$, and $S_\Psi^l$, $S_\Psi^u$, and $S_Y$ are the set of training samples.
$t$ denotes the epoch of training.
$\mathcal{L}_{l}$ and $\mathcal{L}_{u}$ are the loss functions of the supervised and the unsupervised parts, and $\lambda$ is the balance factor for the two losses.

\subsubsection{Supervised Training}
For the CAD representation, there are correlations between the divided parts along the $r$-axis: the closer the prediction is to the label, the smaller the loss of the misclassification should be.
Thus, for the supervised part, we cannot simply use the cross-entropy loss that only considers the label class. We design a multi-stage loss called CADLoss as the supervised loss, formulated as:
\begin{equation}
\label{eq:supervised}
\begin{aligned}
    \mathcal{L}_{l}(\Psi^l, Y, t)
    &= \mathcal{L}_{CAD}(\Psi^l, Y, t) \\
    &= \alpha \mathcal{L}_{dm}(\Psi^l, Y) +
        w_{\textrm{ce}}(t) \mathcal{L}_{ce}(\Psi^l, Y)
    .
\end{aligned}
\end{equation}
The CADLoss contains two parts: mean square error (MSE) with the distance weight of the probabilities, namely the DisMSE loss $\mathcal{L}_{dm}$, and the cross-entropy loss $\mathcal{L}_{ce}$.
$\alpha$ is the balance coefficient.
$w_{ce}(t)$ is the weight controlling the time at which the cross-entropy loss is introduced, expressed as $w_{ce}(t) = \frac{1}{1+e^{-\sigma_1 (t-\mu_1)}}$, where $\sigma_1$ and $\mu_1$ are hyperparameters.
CADNet is first trained under the DisMSE loss to concentrate on the probability distribution of the output around the label, and then the cross-entropy loss is introduced to further increase the probability value of the label class.

The DisMSE and the cross-entropy loss functions are:
\begin{equation}
    \label{eq:dm}
    \mathcal{L}_{dm}(\Psi^l, Y) =
        \frac{1}{N_\varphi}
        \sum_{j=1}^{N_\varphi}
        \sum_{d=1}^{N_r}
        g_{jd} (y_{jd} - \psi^l_{jd})^2
    ,
\end{equation}
\begin{equation}
    \mathcal{L}_{\textrm{ce}}(\Psi^l, Y) =
        \frac{1}{N_\varphi}
        \sum_{j=1}^{N_\varphi}
        \sum_{d=1}^{N_r}
        -y_{jd} \log(\psi^l_{jd})
    ,
\end{equation}
where $j$ and $d$ represent the indexes of angular division and radial division, respectively.
$\psi_{jd} \in [0, 1]$ and $y_{jd} \in \{0, 1\}$ are the prediction probability and label, subject to $\Psi=\{\psi_{jd}|j = 1, ..., N_\varphi, d = 1, ..., N_r\}$ and $Y=\{y_{jd}|j = 1, ..., N_\varphi, d = 1, ..., N_r\}$.
$g$ is the distance weight to reduce the MSE around the label, and $g_{jd} = 1 - e^{-\frac{(d-l_{j})^2}{2\sigma_{g}^2}}, d = 1, ..., N_r$, where $l_j$ is the $j$-th depth ground truth in the index form of label $L=\{l_j|j=1, ..., N_\varphi\}$ and $\sigma_g$ is a hyperparameter.

\begin{table*}[t]
    \caption{Comparison with BEV-based methods on KITTI and SIM}
    \label{tab:bev}
    \centering
    \begin{tabular}{cc|ccccc|ccccc}
        \toprule
              &       & \multicolumn{5}{c}{\textbf{Accuracy} $\uparrow$} & \multicolumn{5}{c}{\textbf{MAE (m)} $\downarrow$} \\
        \textbf{Dataset} & \textbf{Methods} & Total & Thin  & Dynamic & Negative & Others & Total & Thin & Dynamic & Negative & Others \\
        \midrule
        \multirow{6}[0]{*}{KITTI}
            & BEV-M1               & 62.47\%          & 25.15\%          & 23.01\%          & \multirow{5}[0]{*}{-} & 62.92\%          & 0.192          & 0.312          & 0.247          & \multirow{5}[0]{*}{-} & 0.193 \\
            & BEV-M7.5 \cite{MASS} & 56.47\%          & 30.99\%          & 20.98\%          &                       & 56.87\%          & 0.193          & 0.296          & 0.241          &                       & 0.192 \\
            & BEV-M20              & 47.57\%          & 23.39\%          & 23.22\%          &                       & 47.83\%          & 0.188          & 0.272          & 0.262          &                       & 0.187 \\
            & BEV-B                & 40.94\%          & 4.68\%           & 28.92\%          &                       & 41.14\%          & 0.202          & 0.308          & 0.320          &                       & 0.201 \\
            & \textbf{CAD(Ours)}   & \textbf{97.50\%} & \textbf{85.96\%} & \textbf{76.99\%} &                       & \textbf{97.73\%} & \textbf{0.068} & \textbf{0.101} & \textbf{0.120} &                       & \textbf{0.104} \\
        \midrule
        \multirow{6}[0]{*}{SIM}
            & BEV-M1               & 49.99\%          &  6.63\%          &  1.05\%          & 44.07\%               & 67.33\%          & 0.140          & 0.290          & 0.208          & 0.143                 & 0.136 \\
            & BEV-M7.5 \cite{MASS} & 54.79\%          & 61.60\%          & 24.77\%          & 43.51\%               & 71.24\%          & 0.131          & 0.165          & 0.236          & 0.130                 & 0.127 \\
            & BEV-M20              & 52.08\%          & 59.85\%          & 14.60\%          & 42.80\%               & 65.21\%          & 0.134          & 0.172          & 0.247          & 0.130                 & 0.133 \\
            & BEV-B                & 45.99\%          &  5.66\%          &  3.39\%          & 40.33\%               & 62.36\%          & 0.127          & 0.302          & 0.186          & 0.114                 & 0.137 \\
            & \textbf{CAD(Ours)}   & \textbf{97.47\%} & \textbf{92.56\%} & \textbf{92.41\%} & \textbf{98.46\%}      & \textbf{96.83\%} & \textbf{0.050} & \textbf{0.051} & \textbf{0.047} & \textbf{0.051}        & \textbf{0.047} \\
        \bottomrule
    \end{tabular}%
\end{table*}%

\subsubsection{Unsupervised Training}
For the unsupervised part, we make it more concentrated by reducing the variance of the probability distribution.
The variance loss function is:
\begin{equation}
    \mathcal{L}_{var}(\Psi^u) =
        \frac{1}{N_\varphi}
        \sum_{j=1}^{N_\varphi}
        \sum_{d=1}^{N_r}
        (d - \sum_{d=1}^{N_r} d \psi^u_{jd})^2 \psi^u_{jd}.
\end{equation}
However, the probability distribution is relatively uniform at the beginning of the training, resulting in a very large variance, and in the later stage of the training, the variance will become very small. Thus, simply multiplying the variance by a constant factor is not a good choice.
Therefore, we first use entropy regularization \cite{entropy_mini} to obtain an initial probability distribution in the early stage of the training, and then introduce the variance loss to concentrate on the probability distribution in the radial direction.
The entropy regularization loss is:
\begin{equation}
\begin{aligned}
    \mathcal{L}_{reg}(\Psi^u) &=
        \frac{1}{N_\varphi}
        \sum_{j=1}^{N_\varphi}
        \sum_{m=1}^{N_r/b}
        - \psi'_{jm} \log(\psi'_{jm})
        , \\
    \psi'_{jm} &= \sum_{d=b(m-1)+1}^{bm} \psi^u_{jd}
    .
\end{aligned}
\end{equation}
Since we only expect an approximate initial probability distribution, we combine the adjacent $b$ positions to calculate the regularization loss, and $\psi_{jm}'$ is the sum of the adjacent $b$ probability values.

Thus, the total unsupervised loss is:
\begin{equation}
    \label{eq:unsupervised}
    \mathcal{L}_{u}(\Psi^u, t) =
        \mathcal{L}_{reg}(\Psi^u) + w_{var}(t) \beta \mathcal{L}_{var}(\Psi^u)
    ,
\end{equation}
where $\beta$ is a balance factor and $w_{var}(t)$ is responsible for controlling when the variance loss is introduced, expressed as $w_{var}(t) = \frac{1}{1+e^{-\sigma_2 (t-\mu_2)}}$, where $\sigma_2$ and $\mu_2$ are hyperparameters.

\section{Experiments}
In this section, we demonstrate the robustness of traversability prediction delivered through CAD via extensive comparisons and ablation experiments.
We also provide a supplemental video for demonstrating our method on a real UGV and evaluate the feasibility of CAD in real-world scenarios.

\subsection{Perception Experiments}
For CADNet, we set the maximum prediction radius to 15 meters.
Since the spatial feature extraction module has 3 pairs of encoder-decoder structures, the scale of the spatial feature should be a multiple of 8.
We thus set $N_r = 128$ and $N_\varphi=384$ so that the distance resolution and angular resolution are 0.117m and 0.938\degree~respectively, which satisfies the requirements of obstacle avoidance for UGV navigation. 
We choose to retain 4 frames of historical point clouds for multi-frame fusion, i.e. $f=4$. 
The hyperparameters of weights $w_{ce}(t)$ and $w_{var}(t)$ mentioned in Eqs.~(\ref{eq:supervised}) and (\ref{eq:unsupervised}) are set as: $\sigma_1=0.04$, $\mu_1=250$ and $\sigma_2=0.1$, $\mu_2=100$. 
And the two balance factors in Eqs.~(\ref{eq:supervised}) and (\ref{eq:unsupervised}) are set as $\alpha=1$ and $\beta=0.01$.
The hyperparameter $\sigma_g$ of distance weight $g_{jd}$ in Eq.~(\ref{eq:dm}) is set to 9.

The perception experiments include four parts:
\begin{itemize}
    \item Comparison with BEV-based methods in terms of the robustness of traversability prediction.
    \item Ablation experiments of CADNet including the use of different loss functions and different multi-frame fusion methods.
    \item Qualitative analysis of the proposed multi-frame fusion method SAM.
    \item Evaluation of the semi-supervised learning scheme using data collected from multiple scenarios.
\end{itemize}

\subsubsection{Comparison with BEV semantic map}
We first quantitatively compare CAD with BEV semantic map on two datasets. The experimental results are listed in Table \ref{tab:bev}.

\textbf{Dataset.}
Semantic KITTI \cite{KITTI} is one of the most widely used LiDAR point cloud datasets.
However, since its scenarios are for urban autonomous driving, there is a lack of negative obstacles which often occur in UGV navigation.
Thus, we build an environment and collect some data via Isaac Sim simulator \cite{IsaacSim} with a variety of thin, dynamic, and negative obstacles, which more closely resembles the operating scenarios of UGV and better demonstrates the robustness of our method.

Since KITTI's sequence 8 contains a wealth of dynamic objects and can be easily used for comparing the differences among methods, we choose it as the dataset for training and evaluating. KITTI has a total of 4071 samples and SIM has 7932.
Each of them is divided into a training set accounting for 70\% and a validation set accounting for 30\%.
All the experimental results listed in the Table \ref{tab:bev} are produced in the validation sets.

\begin{figure*}[t]
    \centering
    \includegraphics[width=0.95\textwidth]{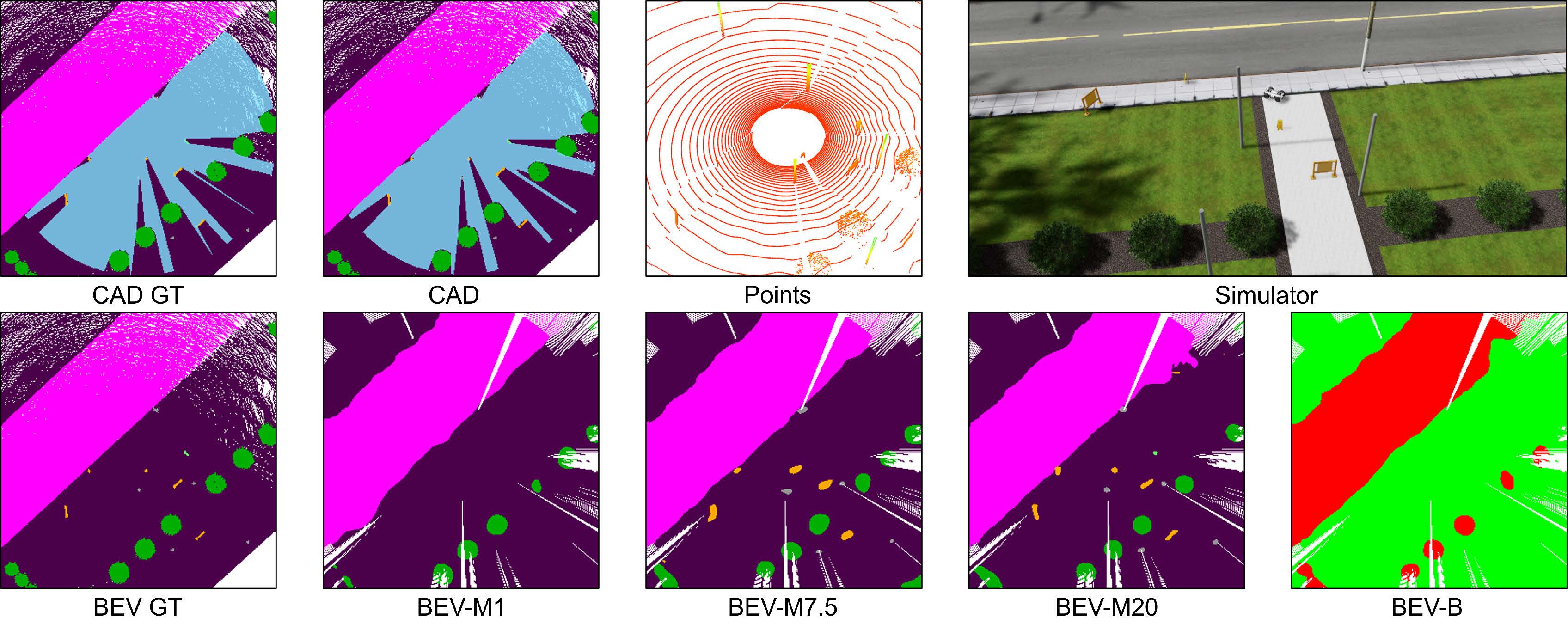}
    \caption{Visual comparison between CAD and BEV semantic map corresponding to thin objects. Top row: the ground truth of CAD, prediction of CAD, point clouds and simulator environment. Bottom row: the ground truth of BEV semantic map, the prediction of BEV-M1, BEV-M7.5, BEV-M20, and BEV-B.
    To represent CAD more clearly, the ground truth and the prediction of CAD (the blue area) are shown together with the BEV semantic map.}
    \label{fig:CADBEV_comp_thin}
\end{figure*}

\textbf{Metrics.}
How to quantitatively compare CAD and BEV semantic map is a thorny issue.
For CAD, the most intuitive evaluation metric is mean absolute error (MAE) of the predicted accessible depths with labels.
However, the ground truth of BEV semantic map is semantic image but not the border distance of the traversable area.

Thus, we propose to compare the two representations by prediction accuracy. For a prediction direction, the predicted traversable distance that differs from the ground truth by no more than 0.5m is considered to be a correct prediction.
We also present the MAE of the predicted accessible depths in those correct prediction directions to illustrate the precision of both representations for predicting the border of traversable area.
Both metrics are presented with 5 situations:
\begin{itemize}
    \item Thin: Objects such as pedestrians, riders, and bicycles in KITTI. In SIM, in addition to pedestrians, a large number of poles and barriers are included.
    \item Dynamic: Moving vehicles, bicycles, and pedestrians.
    \item Negative: Negative obstacles are not included in KITTI. In SIM, the UGV drives on the sidewalk. Thus the downward curb, and the edges of the bridge and pool without guardrails are defined as negative objects.
    \item Others: Other than the above three types of objects, such as static vehicles, sidewalk, and vegetation.
    \item Total: All predicted samples.
\end{itemize}

\textbf{Baselines.}
We choose the open-source BEV-based method MASS \cite{MASS} as the baseline for comparison.
We follow the choice of most BEV methods \cite{semantic_CoRL, MASS} and set the resolution of the BEV semantic map to 0.1m.
With the LiDAR as the origin, the prediction range is 15 meters, i.e., [-15, 15]m for both $x$ and $y$-axis for the Cartesian coordinate-based BEV semantic map, and [0, 15]m for $r$-axis for the polar coordinate-based CAD.
We design multiple experimental configurations for BEV:
\begin{itemize}
    \item BEV semantic map predicts the multi-class semantic information of the environment, while CAD predicts only the accessible depth.
    Therefore, we design a binary-classification BEV to avoid the impact of traversability prediction caused by the number of semantic categories.
    In Table \ref{tab:bev}, the BEV methods with multi-classification are denoted as BEV-M, and the BEV method with binary-classification is denoted as BEV-B.
    \item BEV semantic map is inaccurate in predicting thin objects due to the lack of spatial constraints.
    However, owing to the clear semantic information of categories, BEV semantic map improves the sensitivity of the network to thin objects by setting different category weights.
    For example, MASS \cite{MASS} sets the weights of  pedestrian, bicycle, and rider to 7.5 and those of other categories of objects to 1.
    Therefore, we test the prediction performance of BEV semantic map for thin objects with different semantic weights and compare it with CAD without such pre-defined semantic information.
    We choose 1, 7.5, and 20 as the weights of the thin objects, expressed in Table \ref{tab:bev} as BEV-M1, BEV-M7.5, and BEV-M20, respectively.
\end{itemize}

To make a fair comparison with BEV-based methods, all the CADNet models evaluated in this experiment are obtained only via supervised learning.

\begin{figure*}[t]
    \centering
    \includegraphics[width=0.95\textwidth]{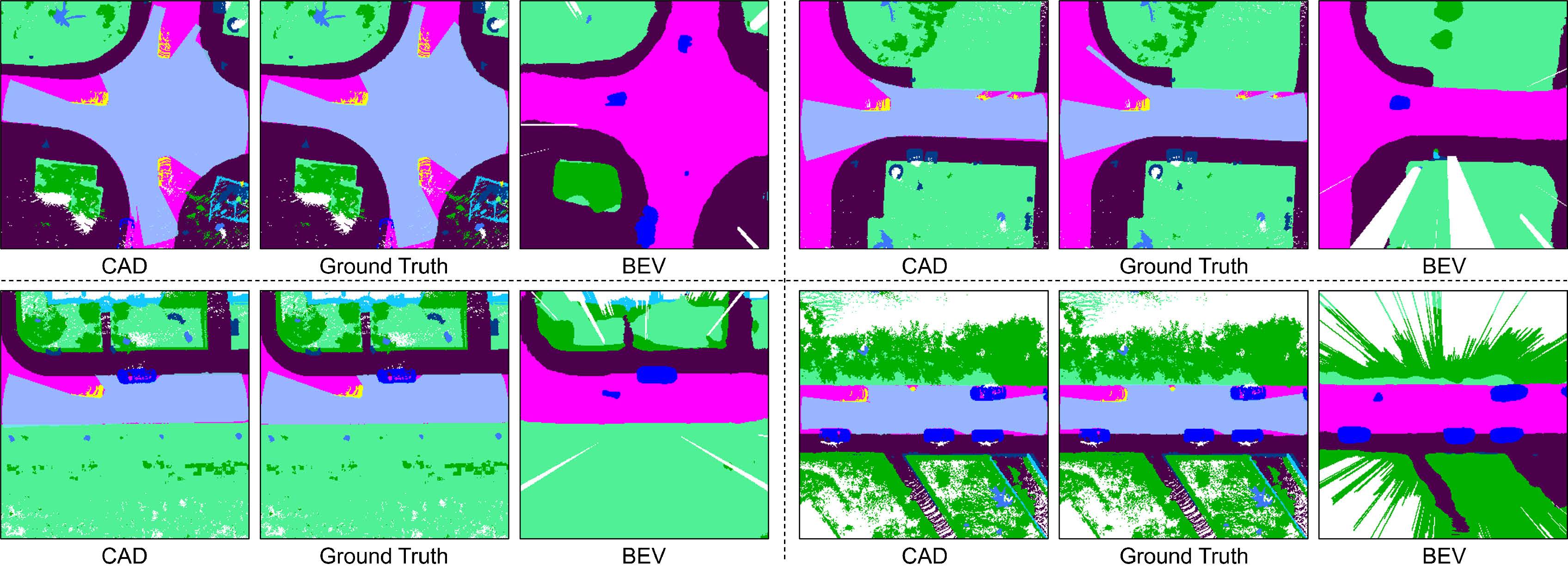}
    \caption{Visual comparison between CAD and BEV semantic map correspoinding to dynamic objects.
    For each of the four parts, the left image is the visualization of CAD prediction, the right one is the visualization of BEV semantic map prediction, and the middle one is the ground truth, which shows the ground truth of both CAD and BEV semantic map.
    To highlight the dynamic objects, they are marked as yellow in the ground truth image. The static and the dynamic objects belonging to the same category have the same labels when leanring BEV semantic maps.
    }
    \label{fig:CADBEV_comp_dyn}
\end{figure*}

\begin{figure*}[t]
    \centering
    \includegraphics[width=0.95\textwidth]{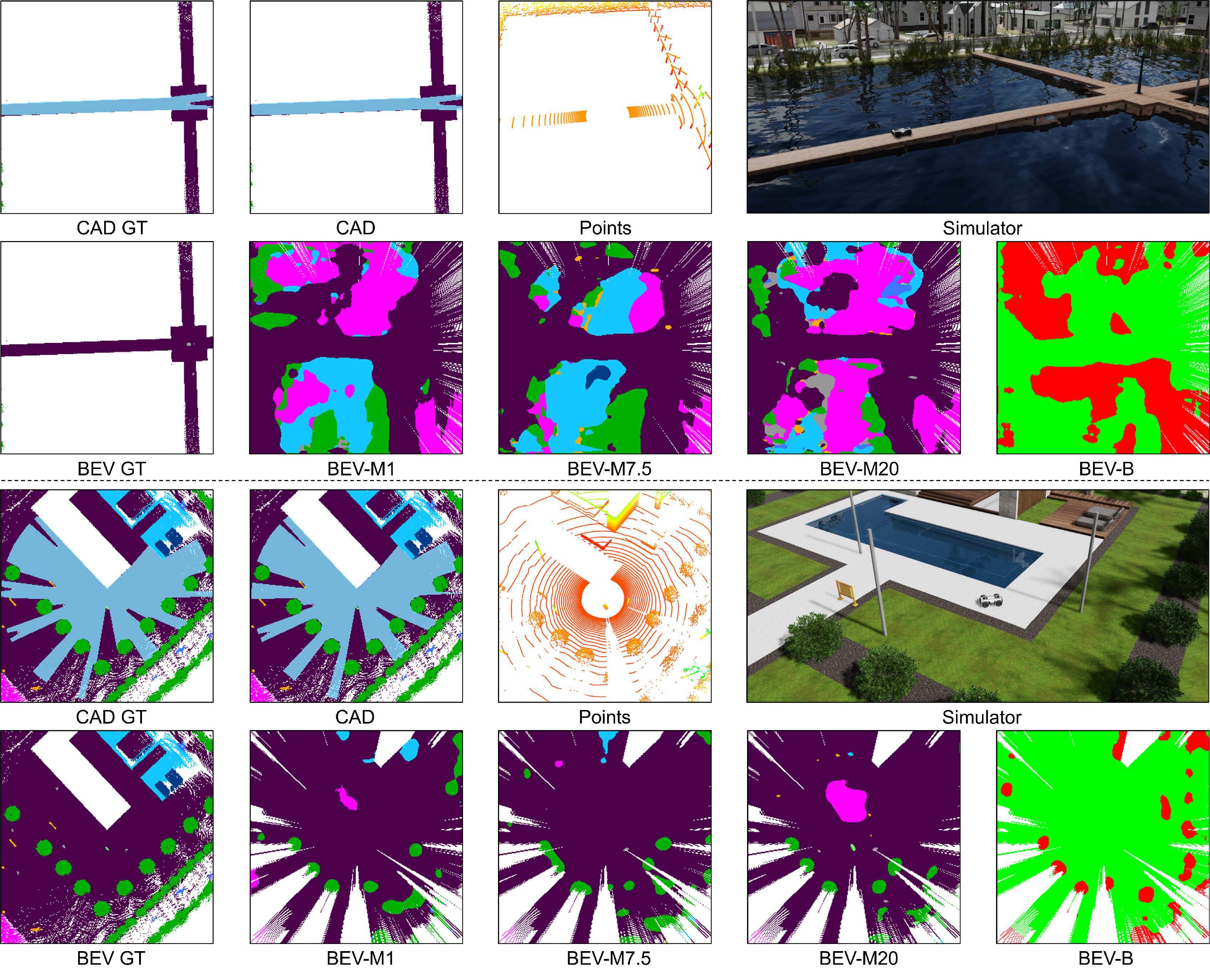}
    \caption{
    Visual comparison between CAD and BEV semantic map corresponding to negative obstacles, including bridge and pool scenarios.
    }
    \label{fig:CADBEV_comp_neg}
\end{figure*}

\textbf{Results.}
As shown in Table \ref{tab:bev}, our method significantly outperforms BEV-based methods on KITTI and SIM.

The prediction accuracy of BEV-M1 with regard to the `Others' category reaches 60\%+ on both datasets while its accuracy with regard to thin objects is very low.
When the weight for thin objects is increased to 7.5, the prediction of thin objects is significantly improved, especially on SIM.
However, when it is increased to 20, the prediction accuracy for thin objects is lower. Besides, we observed that the accuracy for the `Others' category is lower when the weight for thin objects is increasing.
This shows that firstly, there is a limit of improving the prediction accuracy of thin objects just by adjusting the weights, and secondly, blindly increasing the weight of a certain category of objects will deteriorate the prediction of other categories.
Therefore, for BEV-based methods, just adjusting the weights of semantic categories does not fundamentally solve the problem of poor prediction of thin objects.
And for BEV-B, although it seems that the content to be predicted becomes simpler by only binary classification of traversability, the final prediction performance becomes worse due to the lack of pre-defined semantic weights.
By contrast, with the spatial constraints, CAD robustly predicts thin objects with over 90\% accuracy. In particular, CAD significantly outperforms BEV for predicting thin objects in terms of both accuracy and MAE.
The visualization of the prediction results produced by CAD and BEV semantic map is shown in Fig. \ref{fig:CADBEV_comp_thin}.

The annotation of dynamic objects in the ground truth of BEV semantic map is not consistent with the static ones of the same categories, as shown in Fig. \ref{fig:CADBEV_comp_dyn}.
This problem is particularly serious for large objects such as vehicles.
We can see from Table \ref{tab:bev} that BEV semantic map has the lowest prediction accuracy for dynamic objects among all categories of objects.
And it is obvious in the bottom two cases of Fig. \ref{fig:CADBEV_comp_dyn} that BEV semantic map is more accurate in predicting static vehicles, but has a poor performance in predicting the same type of vehicles in motion. In comparison, CAD is more robust in predicting dynamic objects than BEV-based methods.

Negative obstacles can be broadly classified into two types: one is the downward curb when the UGV is on sidewalk, and the LiDAR can detect both the sidewalk and the low ground; the other is the cliff such as the edge of a pool or a bridge without guardrails, and the LiDAR cannot obtain information from the empty locations.
For the first type of negative obstacles, BEV semantic map predicts the semantic categories on both sides of the curb.
However, for the second type of negative obstacles, as shown in Fig. \ref{fig:CADBEV_comp_neg}, due to no effective LiDAR echoes are produced, the BEV ground truth lacks sufficient semantic annotations, which results in a cluttered prediction, and also leads to the low accuracy for predicting the edge locations of negative obstacle.
In comparison, the more reliable ground truth and the introduction of spatial constraints allow the CADNet to focus on predicting the edge locations of negative obstacles, which provides a more robust traversability prediction for UGV to operate in such environments.

\subsubsection{Ablation Experiments}

We evaluate the performance of CADNet with different loss functions and different multi-frame fusion modules to demonstrate the effectiveness of the proposed CADLoss and SAM for CAD prediciton.
The results of the ablation experiments are shown in Table \ref{tab:ablation}, where the data are obtained from the KITTI dataset.
To evaluate the effect of different loss functions, all the models in this experiment are only trained in a supervised manner.

\begin{table}[t]
    \caption{Results of Ablation Experiments\label{tab:ablation}}
    \centering
    \begin{tabular}{c|ccccccc}
        \toprule
        \multirow{2}[0]{*}{\textbf{Methods}} & \multicolumn{2}{c}{\textbf{MAE} $\downarrow$} & \multicolumn{2}{c}{\textbf{Worst-K} $\downarrow$} & \multicolumn{2}{c}{\textbf{IHD} $\downarrow$} \\
                              \cmidrule(lr){2-3}                \cmidrule(lr){4-5}               \cmidrule(lr){6-7}
                              & Total          & Curb           & 5              & 20             & Num.        & Ratio           \\
        \midrule
        L1-Loss               & 0.274          & 0.301          & \textbf{2.801} & 1.588          & -           & -               \\
        CE-Loss               & 0.213          & 0.203          & 4.310          & 2.239          & -           & -               \\
        \midrule
        Single                & 0.177          & 0.187          & 3.052          & 1.334          & -           & -               \\
        Merge \cite{wysiwyg}  & 0.157          & 0.168          & 3.391          & 1.677          & 197         & 6.79\%          \\
        Concat. \cite{MP3}    & 0.169          & 0.177          & 3.603          & 1.838          & 122         & 4.20\%          \\
        \midrule
        \textbf{SAM-5}        & \textbf{0.129} & \textbf{0.130} & 3.025          & \textbf{1.321} & \textbf{32} & 1.10\%          \\
        \textbf{SAM-9}        & \textbf{0.129} & 0.135          & 3.074          & 1.323          & 39          & \textbf{0.98\%} \\
        \bottomrule
    \end{tabular}%
\end{table}%

\textbf{Baselines and metrics.}
The baselines used in the ablation study are divided into two categories.
First, we compare the proposed CADLoss with the commonly used cross-entropy loss and the L1 loss.
Then, we compare several multi-frame fusion methods, including merging all points augmented with timestamp feature \cite{wysiwyg}, concatenating time dimension into channel dimension to fuse temporal feature by 2D CNN \cite{MP3}, and our attention-based SAM. 
We also show the impact of additional frames, cf. SAM-9 in the table, on our method.

We employ three metrics, including MAE, Worst-K, and IHD, to evaluate the performance of different methods:
\begin{itemize}
    \item MAE is the mean absolute error between the output CAD and the ground truth annotation.
    In particular, we show the MAE of the curb prediction to demonstrate the necessity of multi-frame fusion.
    \item Worst-K is the average of the worst K predictions, and we choose K as 5 and 20.
    \item IHD is the interference by historical dynamic objects, as evidenced by two indicators: Number and Ratio.
    The number of IHD represents how many CAD predictions incorrectly point to the location of historical LiDAR points belonging to the dynamic objects.
    And the ratio of IHD represents the percentage of such errors in the prediction direction where the historical LiDAR points appear.
    IHD is used to quantitatively evaluate the immunity of different multi-frame fusion modules to the historical points of dynamic objects.
\end{itemize}

\textbf{Results.}
First of all, for the two loss functions shown in the top two rows of Table \ref{tab:ablation}, the MAE of the classification method trained with the cross-entropy loss is significantly lower than the regression method trained with the L1 loss.
Thus, it is appropriate to consider the task of predicting CAD as a classification task.
However, although the cross-entropy loss for classification outperforms the L1 loss for regression in terms of MAE, it performs not so well in terms of Worst-K.
This is because the cross-entropy loss does not consider the relationship between categories, while in the CAD, neighboring partitions have stronger spatial correlation than distant ones.
Therefore, the CADNet model trained only by cross-entropy outputs some outliers, and suffers from large Worst-K in the quantitative metrics.
For this reason, we propose CADLoss, which concentrates on the probability distribution of CAD around the label.
As shown in the bottom row of Table \ref{tab:ablation}, CADLoss further improves the prediction precision by 40.28\% and significantly reduces Worst-K, which even outperforms the regression method when K is 20.

\begin{figure*}[t]
    \centering
    \includegraphics[width=0.95\textwidth]{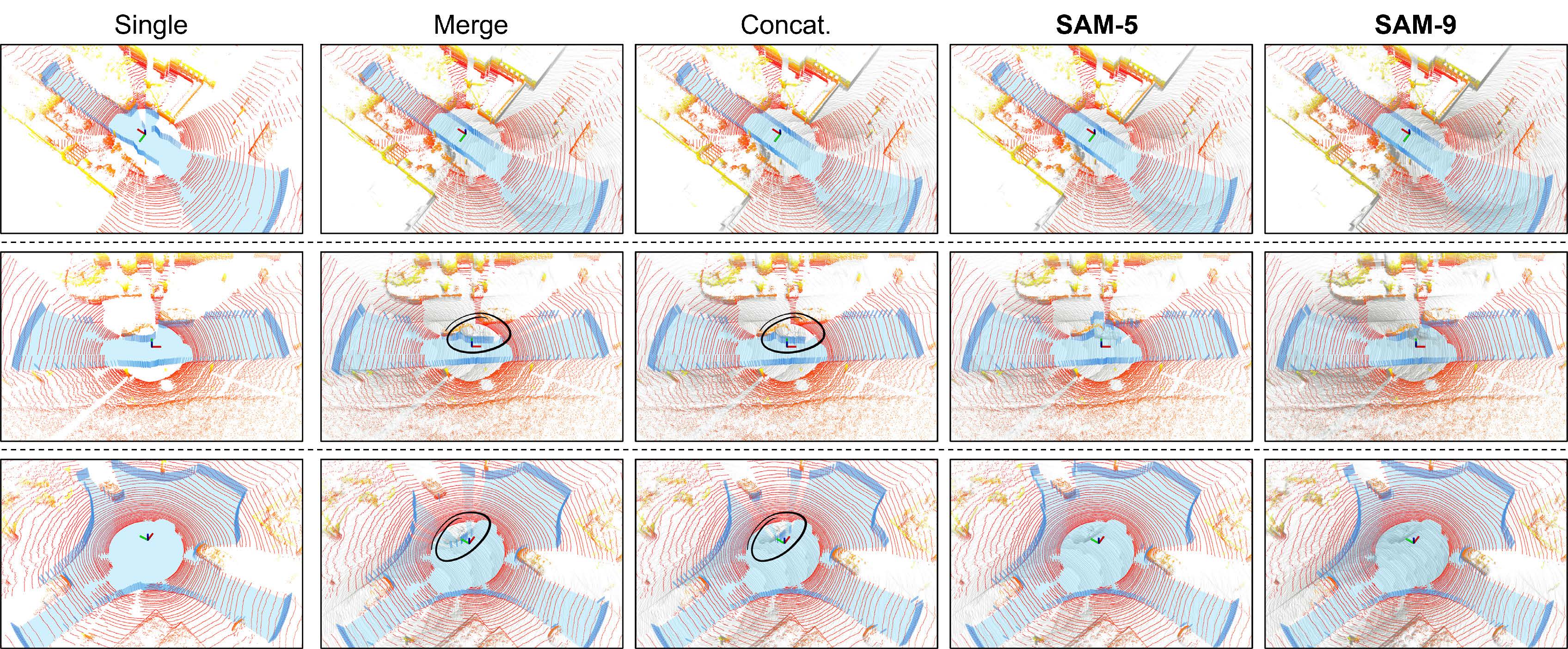}
    \caption{
        Visual comparison of different multi-frame fusion methods.
        The colored points are the points of the current frame, while the gray ones belong to historical frames.
        The blue area represents the predicted CAD.
        The axes in the center of each image represent the position of LiDAR.
    }
    \label{fig:MultiFrameFusion}
\end{figure*}

For the multi-frame fusion, the prediction results of all methods are shown in Fig. \ref{fig:MultiFrameFusion}.
Firstly, as shown in the top left image, due to the large blind spot of LiDAR, a single frame point cloud is not sufficient for accurate prediction, especially for low obstacles like curbs.
Therefore, multi-frame fusion is indispensable for CAD prediction.
The `Merge' and `Concat.' methods can solve the problem that the `Single' method suffers from. However, they are both influenced by the historical points of dynamic objects, cf. IHD in Table \ref{tab:ablation}, and the specific cases are marked in Fig. \ref{fig:MultiFrameFusion} by black circles.
CADNet with the SAM multi-frame fusion structure significantly reduces the interference of dynamic objects.
In addition, we also test the 9-frame version of SAM, and lists the results in the penultimate row of Table \ref{tab:ablation}.
The 9-frame fusion introduces more historical points, resulting in a slight increase in the number of IHD with a lower ratio, which indicates that more frames can increase the stability of SAM.
And as shown in the middle row of Fig. \ref{fig:MultiFrameFusion}, SAM-5 still produces a few wrong predictions, which are corrected by SAM-9.
However, with respect to the efficiency of deployment, the 9-frame version SAM takes nearly twice as long as the 5-frame version SAM.
The number of frames to be fused is related to the number and the density of points in a frame, which needs to be determined empirically, and for KITTI, using 5 frames is a better choice.

\subsubsection{SAM Qualitative Analysis}

In the previous subsection, we quantitatively compare the impact of different multi-frame fusion methods by the historical points of dynamic objects.
In the following, we analyze the working mechanism of SAM in real-world scenarios.

\begin{figure*}[t]
    \centering
    \includegraphics[width=0.95\textwidth]{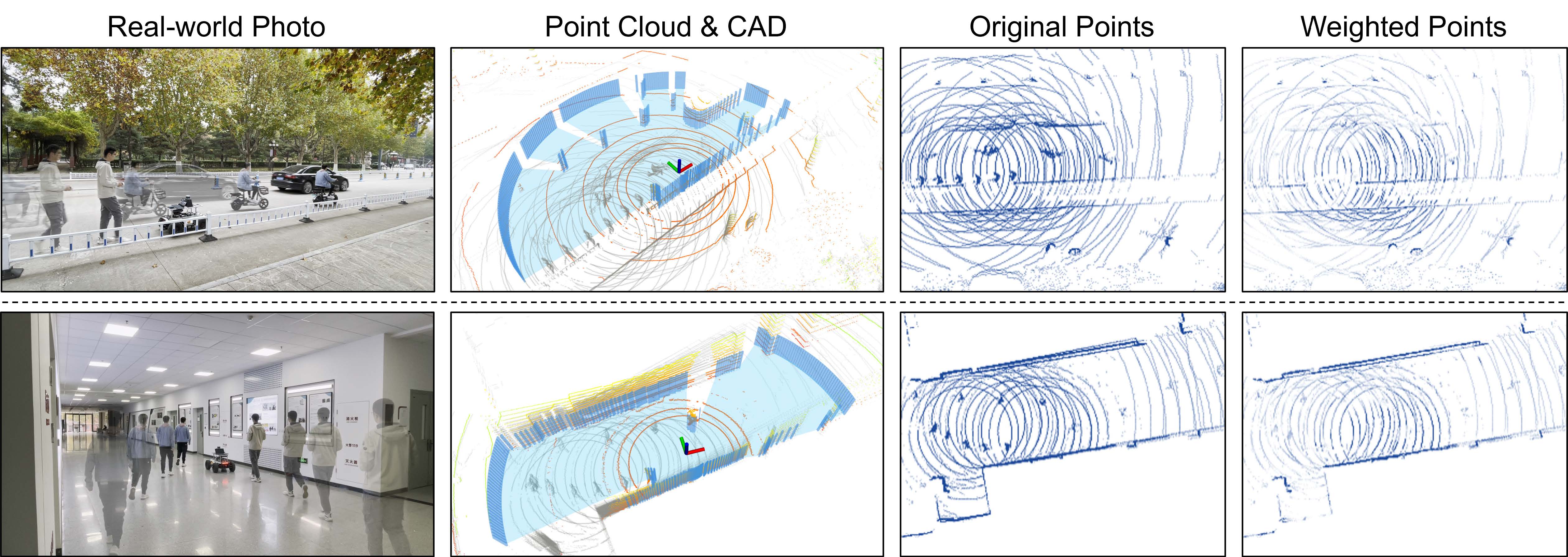}
    \caption{
        Real-world tests of multi-frame fusion.
        From left to right are real-world photos, visualization of predicted CAD with point clouds, top view of the original aggregated multi-frame point clouds, and the top view of the point clouds weighted by the attention weights.
        The two columns of images on the right both contain only the point clouds of historical frames.
    }
    \label{fig:SAM_vis}
\end{figure*}

\begin{figure*}[t]
    \centering
    \subfloat[]
    {
        \includegraphics[width=0.32\textwidth]{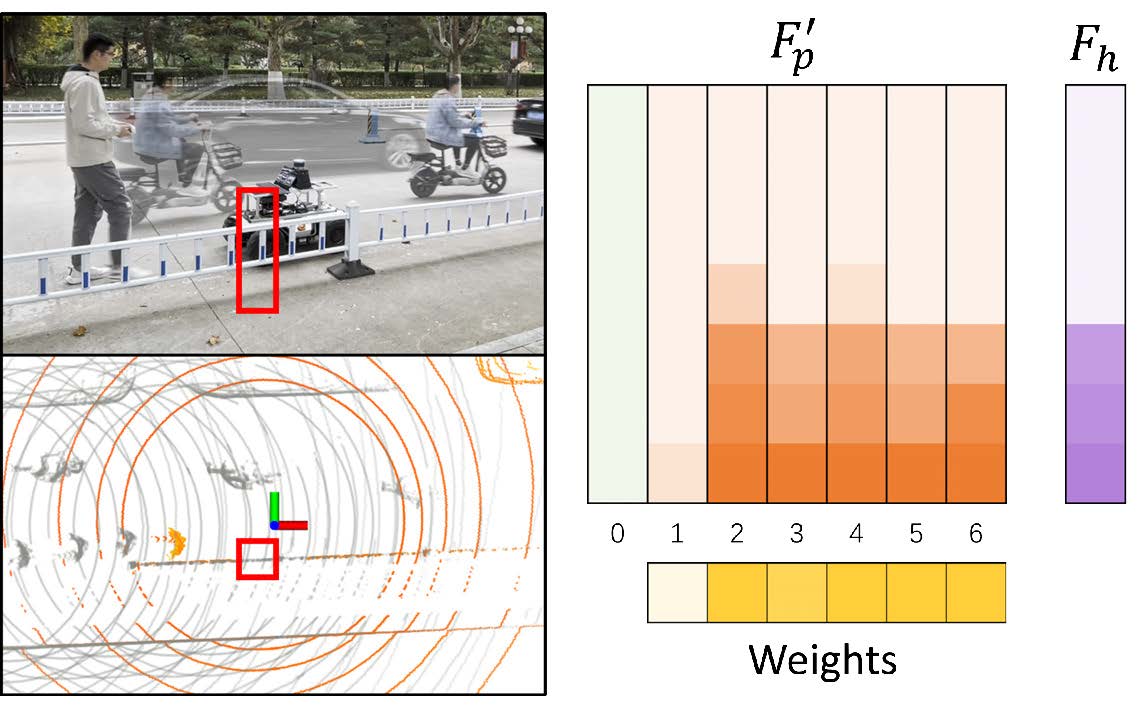}
        \label{fig:SAM_case1}
    }
    \subfloat[]
    {
        \includegraphics[width=0.32\textwidth]{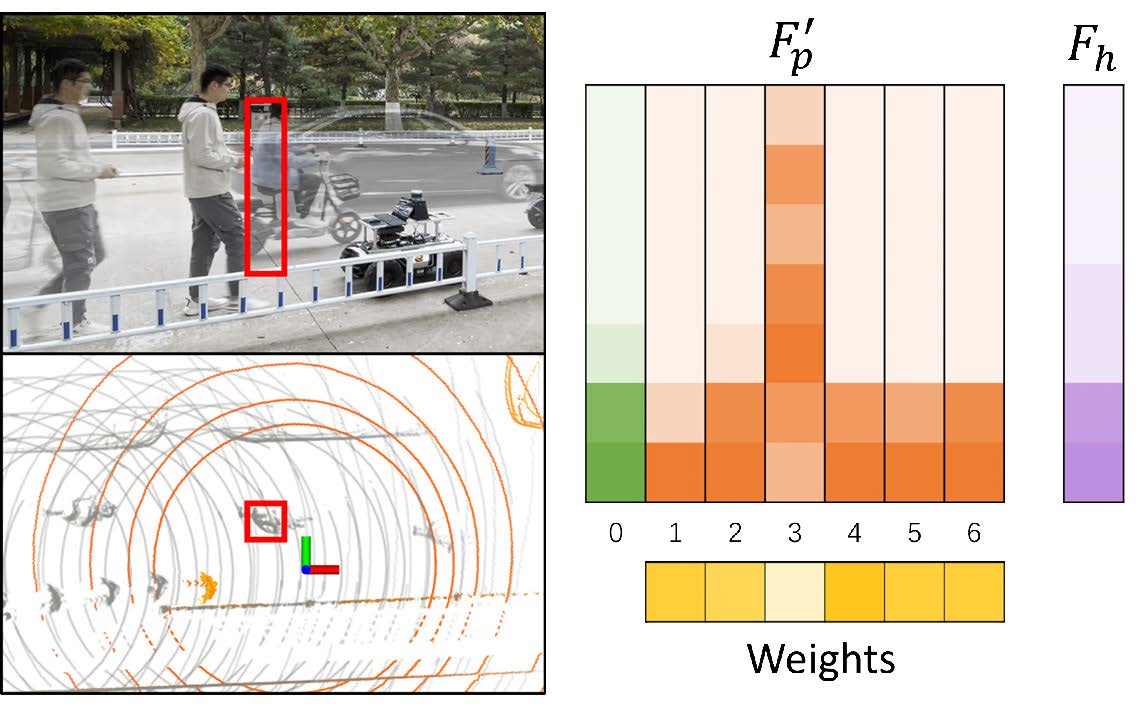}
        \label{fig:SAM_case2}
    }
    \subfloat[]
    {
        \includegraphics[width=0.32\textwidth]{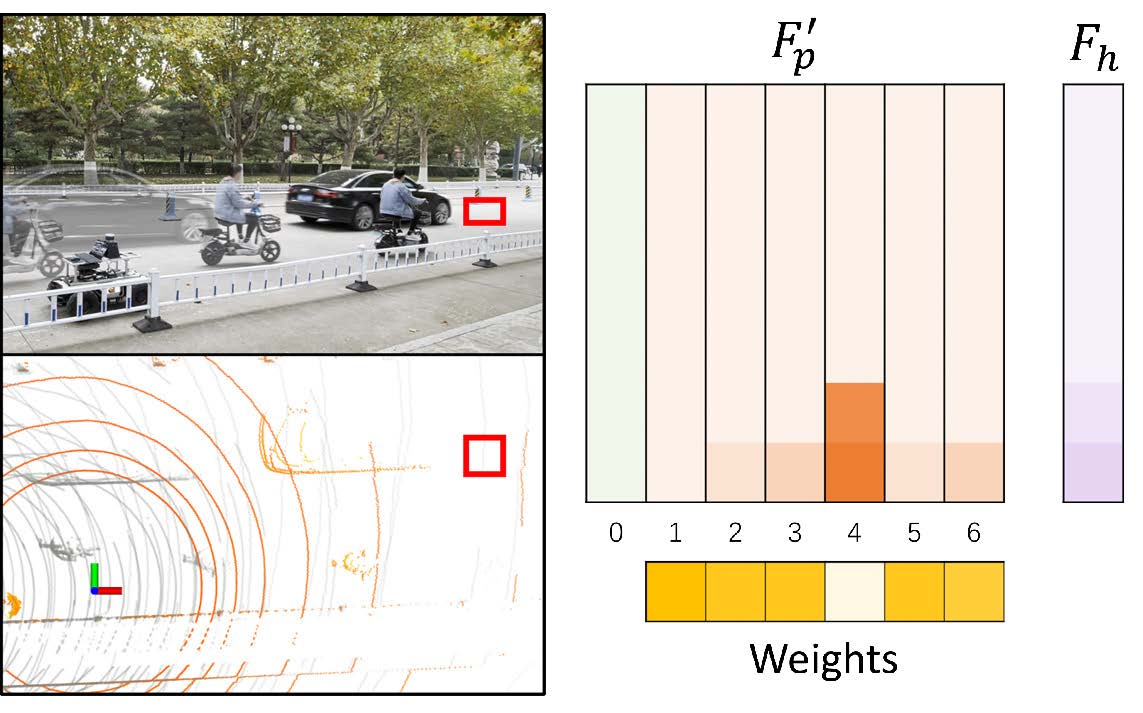}
        \label{fig:SAM_case3}
    }
    \caption{
    Three typical cases of SAM: (a) feature is missing in the current frame; (b) dynamic obstacles present in the historical frame; (c) only individual frames have features.
    In each subfigure, the top left image is the real-world photo, the bottom left one is the point cloud visualization of the corresponding top-down view, and the right image is the illustration of the features and weights in SAM at the location marked with red boxes in the left images.
    }
    \label{fig:SAM_cases}
\end{figure*}

We show the performance of SAM in two scenes containing some dynamic objects as shown in Fig. \ref{fig:SAM_vis}.
We overlap the multi-frame point clouds as shown in the third column. we can observe that the dynamic objects, such as vehicles and pedestrians, lead to trailing shadows in the images.
Then, we multiply each frame of the point cloud image separately by the attention weights generated by SAM and overlap them, as shown in the rightmost column.
It can be seen that most points of dynamic objects are eliminated, and the predicted CAD has no errors at the position of the historical points of such dynamic objects as shown in the second column.

We provide an illustration for each case as shown in Fig. \ref{fig:SAM_cases}, where the two images on the left are the real-world photos and the point cloud visualization.
And the one on the right is the illustration of the features of the spatial location indicated in the left images, where the green and orange parts are the polar features $F_p'$, the yellow row represents the attention weights $w_k, k=1, ..., f$, and the purple column is the fused historical feature $F_h$.
The horizontal direction indicates different frames, and we take $f=6$ as an example. 
The vertical direction indicates the channel of the feature.
The shade of color represents the values of the features or the weights.
We evaluate SAM through the following 3 cases: 

\begin{itemize}
    \item {
    Spatial feature is missing in the current frame but present in the historical ones.
    As shown in Fig. \ref{fig:SAM_case1}, low obstacles, such as curbs and barriers, are in the LiDAR blind spot belonging to this case.
    Although the feature of the current frame is empty, the fused feature is still present as the historical features, which completes the missing spatial feature via multi-frame fusion.
    }
    \item {
    Dynamic object features remain in the history frame.
    This is the main problem that SAM solves as shown in Fig. \ref{fig:SAM_case2}.
    A pedestrian appears in the third historical frame, while all other frames show the feature of ground.
    $F_p^3$ exhibits very different feature from the other frames, thus its weight $w_3$ is small and its feature will not be aggregated into the fused historical feature $F_h$, which enables eliminating the features of dynamic objects.
    }
    \item {
    Only individual frames have features.
    Locations in the distance are only detected by LiDAR at isolated times as shown in Fig. \ref{fig:SAM_case3}.
    In this case, the only feature becomes unstable feature, and therefore is not added into the final fused feature $F_h$, as evidenced by the lightness of the color at the edges in the last column of Fig. \ref{fig:SAM_vis}, which seems to be the opposite of what we expected.
    However, this is actually reasonable because we cannot determine from a single frame whether such a feature is static or dynamic.
    Avoiding fusing such features is the most prudent choice.
    Since this condition usually occurs at a distance, it does not significantly affect CAD prediction.
    }
\end{itemize}

\subsubsection{Evaluation of Semi-supervised Learning}

Semi-supervised learning enables the introduction of a large number of unlabeled samples for training to improve the generalization ability of the model.
We evaluate the improvement of CADNet by semi-supervised training in terms of generalization ability between samples and between scenarios.
We divide the sequences from KITTI into two groups $\mathcal{A}$ and $\mathcal{B}$ based on the scenarios.
Group $\mathcal{A}$ contains sequences 1, 2, 3, 4, 6 and 10, which are highway scenarios with wide roads, and has a total of 9135 samples.
Group $\mathcal{B}$ contains sequences 5, 7, 8 and 9 with more complex scenarios of 9524 samples.
We take one group as the primary group and another as the comparison group, and swap their roles to obtain more comprehensive experimental results.
For the primary group, its samples are randomly divided into three parts: labeled training set, validation set and unlabeled training set, with the proportions of 25\%, 25\% and 50\%, respectively.
And the samples of comparison group are all considered as validation set.

\begin{table}[t]
    \caption{
    Comparison of Semi-supervised Learning Schemes. 
    $P_l$, $P_u$ and $C_u$ represent the labeled training set of the primary group, the unlabeled training set of the primary group and the unlabeled training set of the comparison group, respectively.
    }
    \label{tab:semi}
    \centering
    \begin{tabular}{clcccc}
        \toprule
         \multicolumn{2}{c}{\multirow{2}[0]{*}{\textbf{Training Set}}} & \multicolumn{2}{c}{\textbf{Primary Group}} & \multicolumn{2}{c}{\textbf{Comparison group}} \\
        \cmidrule(lr){3-4} \cmidrule(lr){5-6} 
         && MAE $\downarrow$ & Conf. $\uparrow$ & MAE $\downarrow$ & Conf. $\uparrow$ \\
        \midrule
        \multirow{3}[0]{4em}{PRI $\mathcal{A}$ CMP $\mathcal{B}$}
            & $P_l$  
                & 0.259 & 35.89\% & 0.337 & 32.61\% \\
            & $P_l$+$P_u$ 
                & 0.247 & 62.10\% & 0.308 & 58.68\% \\
            & $P_l$+$P_u$+$C_u$ 
                & 0.246 & 59.94\% & 0.248 & 65.88\% \\
        \midrule
        \multirow{3}[0]{4em}{PRI $\mathcal{B}$ CMP $\mathcal{A}$}
            & $P_l$  
                & 0.174 & 50.40\% & 0.318 & 49.03\% \\
            & $P_l$+$P_u$ 
                & 0.164 & 70.21\% & 0.286 & 69.03\% \\
            & $P_l$+$P_u$+$C_u$ 
                & 0.163 & 58.80\% & 0.266 & 70.80\% \\
        \bottomrule
    \end{tabular}
\end{table}

The experimental results in Table. \ref{tab:semi} are all obtained on the validation sets of both primary and comparison groups.
Firstly, when the training set includes only labeled data, the model has low prediction confidence, and the performance on the comparison group is worse than that on the primary group because of the gap in scenarios between the two groups.
After adding unlabeled samples of primary group $P_u$ for semi-supervised learning, the MAE of the model decreases and the confidence improves for both groups.
The improvement in confidence demonstrates that our semi-supervised loss function (Eq.~\ref{eq:unsupervised}) does concentrate on the output probability distribution of CADNet.
Then, by adding the unlabeled samples of $C_u$ on top of it, the performance of the model improves significantly on the comparison group.
It is worth noting that the training set contains only about 15\% of the labeled samples at this point, which demonstrate that our semi-supervised training procedure effectively improves the generalization ability of CAD with only a small number of labeled samples.

\subsection{Real-world Navigation Experiments}

\begin{table*}[t]
    \centering
    \caption{Quantitative Comparison of Different Navigation Methods through Real-world Experiments.\label{tab:nav}}
    \begin{tabular}{ccccccccccccc}
        \toprule
        \multicolumn{2}{c}{\multirow{2}[0]{*}{\textbf{Senarios}}} & \multirow{2}[0]{*}{\textbf{Methods}} & \multicolumn{2}{c}{\textbf{Success Rate}} & \multicolumn{2}{c}{\textbf{Collision-Ped}} & \multicolumn{2}{c}{\textbf{Collision-Neg}} & \multicolumn{2}{c}{\textbf{Collision-Other}} & \multicolumn{2}{c}{\textbf{Avg. Speed}} \\
        \multicolumn{2}{c}{} & & Static & Dynamic & Static & Dynamic & Static & Dynamic & Static & Dynamic & Static & Dynamic \\
        \midrule
        \multirow{6}[0]{*}{Outdoor}
            & \multirow{3}[0]{*}{Driveway}
                & Auto. \cite{autoware} & 83.3\% & 50.0\% & 2 / 2  & 4 / 6  & -      & -      & 0 / 2   & 2 / 6  & 0.91 & 0.80 \\
            &   & STVL \cite{STVL}      & 66.7\% & 50.0\% & 0 / 4  & 2 / 6  & -      & -      & 4 / 4   & 4 / 6  & 0.79 & 0.72 \\
            &   & \textbf{Ours}         & 100\%  & 100\%  & 0 / 0  & 0 / 0  & -      & -      & 0 / 0   & 0 / 0  & 0.95 & 0.82 \\
        \cmidrule{2-13}
            & \multirow{3}[0]{*}{Sidewalk}
                & Auto. \cite{autoware} & 0.00\% & 0.00\% & 0 / 12 & 3 / 12 & 0 / 12 & 0 / 12 & 12 / 12 & 9 / 12 &  -   &  -   \\
            &   & STVL \cite{STVL}      & 66.7\% & 16.7\% & 0 / 4  & 1 / 10 & 4 / 4  & 9 / 10 &  0 / 4  & 0 / 10 & 0.84 & 0.67 \\
            &   & \textbf{Ours}         & 100\%  & 100\%  & 0 / 0  & 0 / 0  & 0 / 0  & 0 / 0  &  0 / 0  & 0 / 0  & 0.90 & 0.78 \\
        \midrule
        \multirow{6}[0]{*}{Indoor}
            & \multirow{3}[0]{*}{Corridor}
                & Auto. \cite{autoware} & 83.3\% & 33.3\% & 0 / 2 & 5 / 8 & - & - & 2 / 2 & 3 / 8 & 0.91 & 0.82 \\
            &   & STVL \cite{STVL}      & 100\%  & 66.7\% & 0 / 0 & 4 / 4 & - & - & 0 / 0 & 0 / 4 & 0.85 & 0.65 \\
            &   & \textbf{Ours}         & 100\%  & 100\%  & 0 / 0 & 0 / 0 & - & - & 0 / 0 & 0 / 0 & 0.99 & 0.88 \\
        \cmidrule{2-13}
            & \multirow{3}[0]{*}{Stairhall}
                & Auto. \cite{autoware} & 83.3\% & 58.3\% & 2 / 2 & 2 / 5 & 0 / 2 & 0 / 5 & 0 / 2 & 3 / 5 & 0.89 & 0.85 \\
            &   & STVL \cite{STVL}      & 58.3\% & 50.0\% & 0 / 5 & 0 / 6 & 4 / 5 & 4 / 6 & 1 / 5 & 2 / 6 & 0.76 & 0.63 \\
            &   & \textbf{Ours}         & 100\%  & 100\%  & 0 / 0 & 0 / 0 & 0 / 0 & 0 / 0 & 0 / 0 & 0 / 0 & 0.95 & 0.84 \\
        \bottomrule
    \end{tabular}%
\end{table*}%

To demonstrate the feasibility of our method in real-world applications, we deploy CAD on a real UGV and compare it with several UGV navigation approaches.

\subsubsection{Setups}
We deploy our method on an AgileX SCOUT UGV, equipped with WTGAHRS1 IMU, TOP103 GPS, and Velodyne VLP-16 LiDAR. 
LiDAR is mounted above the center of the UGV at a height of 0.8m from the ground. 
The program runs on an on-board computer with an i5-9400H CPU and a GTX-1650 GPU. 
It spends about 60ms handling a frame, which satisfies the real-time requirement.
We implement the Timed-Elastic-Band (TEB) \cite{TEB} local planner as the downstream local planner for CAD.

\textbf{Scenarios.}
We set up 4 real scenarios, including the outdoor ones `Driveway' and `Sidewalk', and the indoor ones `Stairhall' and `Corridor', to quantitatively evaluate our approach and the baselines. 
Among them, Sidewalk and Stairhall contain negative obstacles such as downward curbs and stairs.
Each competing method was experimented 12 times with static and dynamic obstacles respectively in each scenario.

\textbf{Metrics.}
We evaluate the UGV's navigation performance using three metrics: Success Rate, Collision, and Average Speed. Success Rate is the ratio of the number of times the UGV reaches the navigation goal without collision. Average Speed is the average speed of the UGV in all successful navigation tests.
We divide the collision situation into three categories according to the objects that produce collisions, which are pedestrians, negative obstacles and others.

\textbf{Baselines.}
We chose two widely used and open-source autonomous navigation approaches, STVL \cite{STVL} and Autoware \cite{autoware}, as baselines for comparison. 
STVL is a voxel-based method, generating 2D costmap based on the voxel grid for the planning module. 
We implemented the original TEB local planner with it to complete navigation experiments. 
Then we tested the Autoware \cite{autoware} where the perception module is based on the popular PointPillars \cite{PointPillars}.

\subsubsection{Results and Analysis}
The quantitative results are listed in Table~\ref{tab:nav}, and please refer to the supplemental video for the practical navigation performance.

STVL produces a 2D costmap by projecting the LiDAR points within a pre-set height range into a 2D map.
This way of calculating obstacle positions through point cloud geometric relationships has relatively high stability, while the pre-set height threshold poses some problems for practical applications.
If the threshold is too low, the LiDAR points on the ground will be also considered as obstacles, making the UGV unable to drive properly.
And if the threshold is too high, the UGV will ignore some low obstacles and collide with them.
From the Collision-Other column of Table \ref{tab:nav}, it can be seen that STVL has multiple-collisions in the Driveway scenario, and most of them are with curbs.
In addition, STVL cannot handle the negative obstacles such as downward curbs in the Sidewalk scenario and the downward stairs in the Stairhall scenario.

Autoware is an comprehensive autonomous driving framework that works on high-definition maps, where the locations of curbs and stairs are pre-defined.
The 3D detection-based Autoware cannot detect such obstacles either, and thus it works around the issue of negative obstacles by high precision positioning combined with high-definition maps.
And in the event of a positioning deviation, it may still lead to a collision.
Also, 3D detection cannot recognize such objects as trash cans, lamp posts, and tree trunks in the Sidewalk scenario, and tables, chairs, and shelves in the Corridor scenario, resulting in a number of collisions.

By comparison, our method based on CAD does not require to set fixed thresholds or rely on high definition maps and can be simply applied to UGV navigation.
This is because CAD uniformly represents various obstacles, including vehicles, pedestrians, curbs, and negative obstacles, and thus provides a robust traversability prediction in different environments.
In the test scenarios, our method did not collide with any object and achieved the highest success rate.

\section{Conclusion}
We propose a robust traversability representation, namely Circular Accessible Depth (CAD), to represent the road traversability for UGV navigation. 
We design a CADNet with an attention-based multi-frame fusion module SAM to extract the spatial features of the LiDAR points for CAD prediction. 
The CAD representation can be well applied to a semi-supervised learning scheme with easily acquired ground truth, which significantly facilitates its deployment in practice. 
We demonstrate through perception and real-world experiments that CAD is more robust in traversability prediction than the existing BEV semantic maps, Autoware and STVL for UGV navigation. 
Our current design for CAD is based on the premise that traversability is binary, while this limits the deployment range of CAD.
Thus, in the future, we plan to design a multi-level CAD to represent the traversability more appropriately in a wider range of scenarios.


\bibliographystyle{IEEEtran}
\bibliography{reference}

\end{document}